\documentclass{article}
\usepackage{spconf,amsmath,graphicx}
\usepackage{cite}
\usepackage{booktabs}
\usepackage{amssymb}
\usepackage{multicol, blindtext}
\usepackage{caption}
\usepackage{multirow}
\usepackage{xcolor}
\usepackage{url}

\def\x{{\mathbf x}}

\title{Hierarchical Attention Fusion for Geo-Localization}
\name{Liqi Yan$^{1}$, Yiming Cui$^{2}$, Yingjie Chen$^{3}$, Dongfang Liu$^{3*}$\thanks{$^{*}$Dongfang Liu is the corresponding author.}}
\address{
{$^{1}$Department of Computer Science and Technology, Fudan University}\\
{$^{2}$Department of Electrical and Computer Engineering, University of Florida} \\
{$^{3}$Department of Computer Graphics Technology, Purdue University} 
}
\begin{document}
%\ninept
%
\maketitle
\begin{abstract}
Geo-localization is a critical task in computer vision. In this work, we cast the geo-localization as a 2D image retrieval task. Current state-of-the-art methods for 2D geo-localization are not robust to locate a scene with drastic scale variations because they only exploit features from one semantic level for image representations. To address this limitation, we introduce a hierarchical attention fusion network using multi-scale features for geo-localization. We extract the hierarchical feature maps from a convolutional neural network (CNN) and organically fuse the extracted features for image representations. Our training is self-supervised using adaptive weights to control the attention of feature emphasis from each hierarchical level. Evaluation results on the image retrieval and the large-scale geo-localization benchmarks indicate that our method outperforms the existing state-of-the-art methods. Code is available here: \url{https://github.com/YanLiqi/HAF}.
\end{abstract}
\section{Introduction}
Geo-localization is an important task in computer vision as it holds valuable potentials for applications such as autonomous driving \cite{9206716} and robot navigation \cite{9207265}. When working under the region with poor GPS signals, mobile agents require a supplementary localization for operation and geo-localization is a helpful addition to GPS \cite{liu2020visual}.

In our work, we cast the geo-localization problem as a task of image retrieval \cite{arandjelovic2016netvlad}, which searches over a pre-stored GPS-tagged image database to determine the current  location \cite{salarian2018improved}. The GPS-tagged image from the database with the closest distance to the query image in feature space is approximated as the location of the query \cite{sarlin2019coarse}. 
\begin{figure}[!ht]
  \centering
\includegraphics[width=7cm]{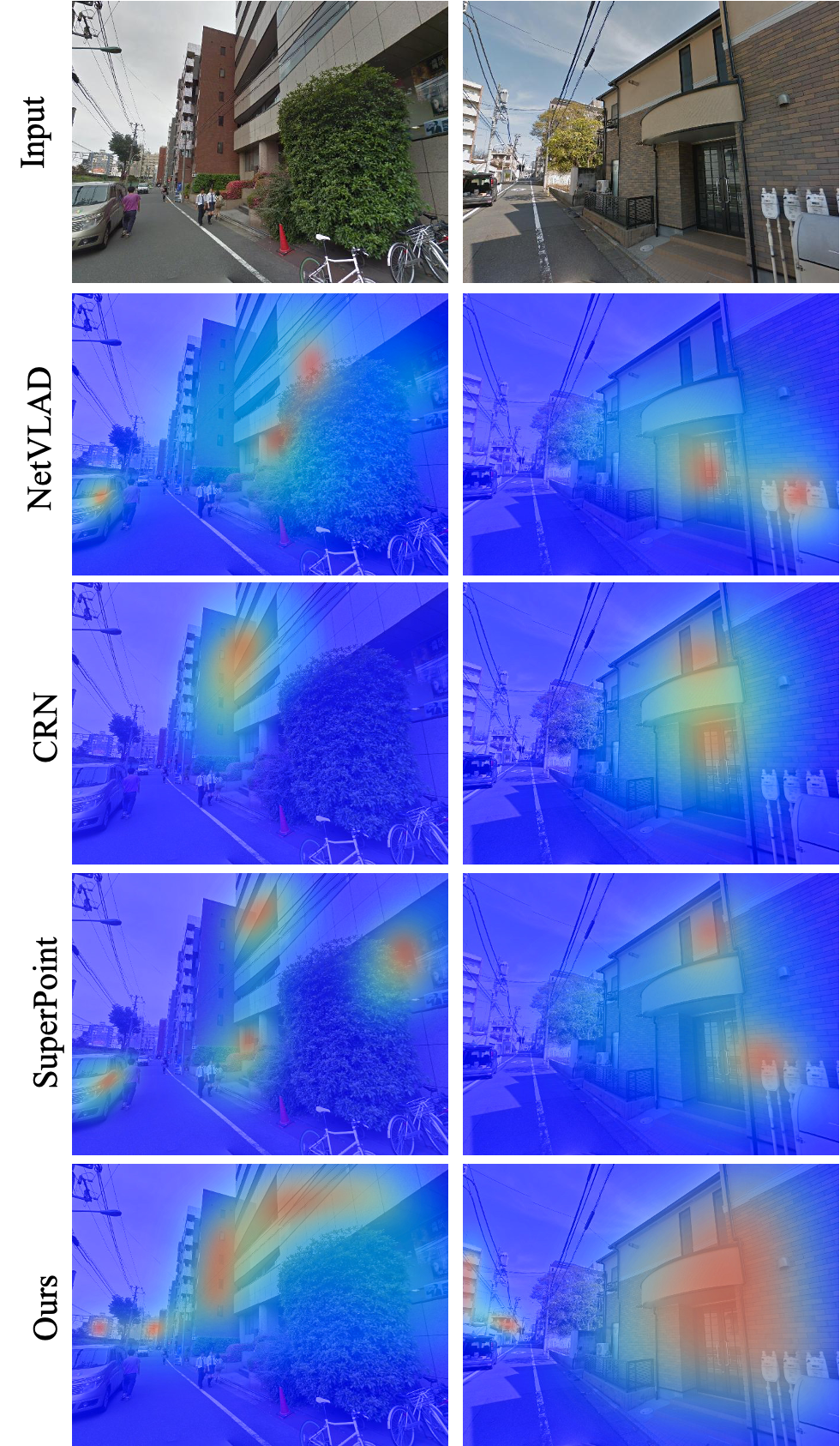}
  \vskip -6pt
  \caption{Comparison of feature emphasis. Compared to conventional methods~\cite{arandjelovic2016netvlad,jin2017learned,detone2018superpoint}, our method exploits the multi-scale features for hierarchical attention to depict image representation of landmarks with different scales and distance.}
  \label{Figure 1}
    \vskip -12pt
\end{figure}
\subsection{Related Work}
Nowadays, convolutional neural networks (CNNs) have become a  powerful technique to explore image representations \cite{simonyan2014very,LIU20201}. The primary challenge for geo-localization is to produce discriminative image representation to identify two different places \cite{salarian2018improved,liu2020visual}. A geo-localization database generally contains images having landmarks with different scales. For conventional methods~\cite{arandjelovic2016netvlad,jin2017learned,detone2018superpoint}, landmarks, with medium or small sizes, are difficult to be recognized because CNNs intend to downsample the spatial resolution of the input image by a significant margin \cite{lin2017feature,huang2017densely}.  However, many medium- and small-size landmarks include valuable  distinctiveness in image representation for geo-localization \cite{jin2017learned}. Thus, our method wants to exploit multi-scale features to delineate landmarks with different distances and scales (Fig. \ref{Figure 1}).

A critical reason for the aforementioned problem is that concurrent methods~\cite{arandjelovic2016netvlad,sarlin2019coarse,jin2017learned,detone2018superpoint} only use features from one semantic level for the geo-localization task. The feature maps from a single semantic level fail to fully explore rich visual clues from landmarks of different scales. This observation motivates us to exploit hierarchical features with different semantics to improve the geo-localization task.

\begin{figure*}[!ht]
  \centering
\includegraphics[width=13cm]{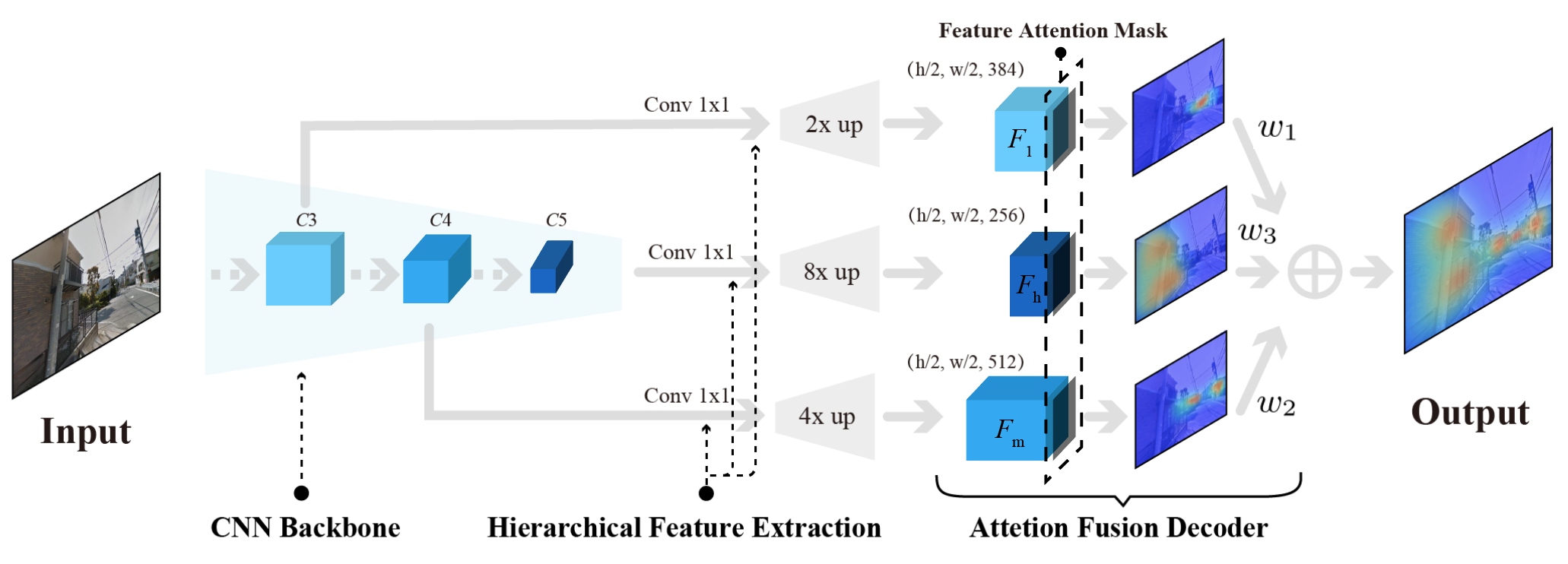}
  \vskip -6pt
  \caption{The architecture of the proposed method. Our method uses hierarchical features to  close  the  semantic gap  in  feature  learning. We perform the attention fusion over the obtained features to produce strong image representation for landmarks with different scales.}
  \label{Figure 2}
  \vskip -12pt
\end{figure*}
\subsection{Principal Contributions}
Our work brings the following three contributions. First, we introduce a hierarchical attention fusion network, a novel algorithm for geo-localization (Fig. \ref{Figure 2}.). We find inspiration from the feature pyramid networks \cite{lin2017feature} which uses hierarchical feature maps to predict multi-scale image representations. Different from the conventional methods\cite{arandjelovic2016netvlad,detone2018superpoint}, we extract hierarchical features at the lower-level, mid-level, and higher-level layer of the backbone network. Thus, the obtained features have rich multi-scale information. We then perform a feature fusion over the obtained features, so that our method can simultaneously pay attention to landmarks with various scales to identify discriminative keypoints for geo-localization. Second, we propose a self-supervised loss function to captures pairwise image relationships in training. Our training only needs GPS-tagged image pairs instead of expensive pixel-wise correspondences between images. Our training strategy can encourage the trained algorithm to learn which imagery context should be focused or suppressed to achieve a better image representation for localization. Finally, we use extensive experiments to assess our method. Results demonstrate that the proposed method sets a  new state-of-the-art on several geo-localization benchmarks.
\section{Method}
This section details our method. The architecture of our method is shown in Fig. \ref{Figure 2}, which includes two principal modules: (i). Hierarchical feature extraction, and (ii). Feature fusion decoder. Using CNN, we first extract hierarchical feature maps to close the semantic gap in feature learning. We then perform attention fusion over the obtained multi-scale features to predict strong image representation.
\subsection{Hierarchical Feature Extraction}
We use VGG16 \cite{simonyan2014very} as the backbone network for feature extraction. We extract hierarchical features from  \texttt{Con3\_2}, \texttt{Con4\_3}, and \texttt{Con5\_3} respectively (Fig. \ref{Figure 2}.).  The corresponding feature maps are $\{C3,C4,C5\}$ and each has strides of $\{4, 8, 16\}$ pixels with respect to the input image. The obtained hierarchical feature maps are then processed by a modified SuperPoint structure \cite{detone2018superpoint}, which includes a non-linear $1\times1$ convolutional layer to control the channel dimensions and an upsampling layer to increase the feature map resolution. Specifically, the non-linear $1\times1$ convolutional layer uses \texttt{ReLU6} to determine the feature activation. The upsampling layer increases the resolution of the feature maps in a non-learned manner. In order to reduce the aliasing effect from upsampling, we perform the Adaptive Spatial Fusion on each output  \cite{guo2020augfpn}. The output set of feature maps is $\{F_{l}, F_{m}, F_{h}\}$ corresponding to $\{C3, C4, C5\}$ as lower-level, mid-level, and higher-level features. Compared to $\{C3, C4, C5\}$, the channel dimensions for $\{F_{l}, F_{m}, F_{h}\}$ are multiplied 1.5$\times$,  1$\times$,  0.5$\times$  times  respectively  and their  spatial  resolutions  is  brought  back  to  half  of the  input  resolution. $\{F_{l}, F_{m}, F_{h}\}$ together  has rich features of 1,152 dimensions.
\subsection{Attention Fusion Decoder}
\textbf{Feature attention mask.} We denote the channel dimensions for $F_{l}, F_{m}, F_{h}$ as $x$, $y$, and $z$ respectively. Thus, $F_{l}, F_{m}, F_{h}$  can be expressed as a set of feature representations as $\{f_{1},... , f_{x}\}$, $\{f_{1},... , f_{y}\}$, and $\{f_{1},... , f_{z}\}$ accordingly, where $f$ is one feature map from a single channel.

Inspired  by \cite{jin2017learned}, we implement three learnable feature attention masks $\{m_1, m_2, m_3\}$ which are appended to $F_{l}, F_{m}, F_{h}$ separately. The attention mask is to indicate which spatial regions from the feature maps are discriminatively representative for localization. We define the attention-weighted features as:
\begin{equation}
% \scriptsize
% \small
\footnotesize
% \begin{aligned}
%     F'_{l} &=\sum_{n=1}^{x}{\sum_{r\in R}m^{r}\cdot{f^{r}_{n}}} \\
%     F'_{m} &=\sum_{n=1}^{y}{\sum_{r\in R}m^{r}\cdot{f^{r}_{n}}} \\
%     F'_{h} &=\sum_{n=1}^{z}{\sum_{r\in R}m^{r}\cdot{f^{r}_{n}}} \\
% \end{aligned}
    F'_{l} =\sum_{n=1}^{x}{\sum_{r\in R}m^{r}\cdot{f^{r}_{n}}},
    F'_{m} =\sum_{n=1}^{y}{\sum_{r\in R}m^{r}\cdot{f^{r}_{n}}},
    F'_{h} =\sum_{n=1}^{z}{\sum_{r\in R}m^{r}\cdot{f^{r}_{n}}} 
\label{Equation 1}
\end{equation}
where $R$ denotes a set of spatial regions on the feature map. During backpropagation, each $m$ is learned to emphasize or suppress certain features from different spatial regions to encourage discriminative representation.

 \textbf{Coupled descriptor and  detector.} Similar to \cite{detone2018superpoint}, we implement a coupled detector and descriptor based on the weighted features to prevent information loss. Using the attention-weighted features $F'$, we define descriptor as a set of  vectors $K$:
\begin{equation}
% \scriptsize
\small
    K=\sum_{i=1}^{h}\sum_{j=1}^{w}F'^{ij:}, K^{ij} \in \mathbb{R}^{x}.
    \label{Equation 2}
\end{equation}
The obtained descriptors $K$ are ${L}_{2}$-normalized  to be a unit length. At a pixel point ($i, j$) on the image, we calculate the Euclidean distance of each descriptor $K^{ij}$ between images to establish feature correspondences.
 For detectors, we exploit the feature maps $F'$ in the same manner. Thus the detectors $D$ can be denoted as:
\begin{equation}
% \scriptsize
\small
    D =\sum_{n=1}^{x} {F'}^{::n}, \quad {D}^{n} \in {\mathbb{R}}^{h\times w}
    \label{Equation 3}
\end{equation}
If the pixel point $(i, j)$ is detected, we denote the most strong detection of all channels as $D^{(ij)n'}$ ($n' \in \mathbb{R}^{x}$) on the response maps. We then  perform an image-wise normalization of the  detection to obtain the detection score at a pixel $(i,j)$:
\begin{equation}
% \scriptsize
\small
    {s}_{ij} = \frac{D_l^{(ij)n'}}{\sum_{i'=1}^{h}{\sum_{j'=1}^{w}{D_l^{(ij)n'}}}}.
    \label{Equation 4}
\end{equation}
We perform attention fusion decoding by using all the detectors and descriptors from different hierarchical levels to predict the image representation (Fig. \ref{Figure 2}).
\subsection{Training Objective}
All our modules are trained in an end-to-end fashion which facilitates the task-relevant feature learning. Given a query image $I_q$, our goal is to approximate its location by finding the reference images $\{I_r\}$  which are the nearest neighbors in feature space. To achieve hierarchical attention fusion, we propose a novel triplet ranking loss to jointly optimize the detectors and descriptors based on the different hierarchical features. Our training is weakly supervised. Thus, instead of using expensive feature correspondences at the pixel level to learn, our training only needs image-level annotations (the positive references $\{I^+_r\}$ and the negative references $\{I^-_r\}$). Namely, our method is trained to match the positive references $\{I^+_r\}$ and discriminate the negative ones $\{I^-_r\}$.

For a pair of image $({I}_{q}, {I}_{r})$, we define their feature differences by calculating the the descriptor distance $\sum_{c \in \mathcal{C}}\parallel {K}^{c}_{q} - {K}^{c}_{r} \parallel_{2},$ where $\mathcal{C}$ indicates all the  corresponding feature  points between the two images. In training, we maximize the distance of the corresponding descriptors between the negative pairs while minimizing the distance between the positive ones. Additionally, in order to increase the detection repeatability \cite{dusmanu2019d2}, we  include a detection term to compute differences in feature  space between two images:

\begin{equation}
% \scriptsize
\small
    \Delta\mathcal{D}\left(I_{q}, I_{r}\right) =  \sum_{c \in \mathcal{C}}\frac{s_{q}^{c}s_{r}^{c}}{\sum_{c’ \in \mathcal{C}}s_{q}^{c'}s_{r}^{c'}} \parallel {K}^{c}_{q} - {K}^{c}_{r} \parallel_{2}.
    \label{Equation 5}
\end{equation}
where $s$ is the detection scores in (4). Thus, the triple tranking loss is defined as:
\begin{equation}
% \scriptsize
\footnotesize
% \small
\mathcal{L}\left(I_q, I^+_r, I^-_r\right)=\max\left(M + \Delta\mathcal{\mathcal{D}}\left(I_{q}, I^+_{r}\right) - \Delta\mathcal{D}\left(I_{q}, I^-_{r}\right),0\right)
\label{Equation 6}.
\end{equation}
Since we jointly optimize the detectors and descriptors based  on  the  different  hierarchical  features, our overall loss is:  
\begin{equation}
% \scriptsize
\small
\mathcal{L}_{total}=w_1\cdot\mathcal{L}_{l}+w_2\cdot\mathcal{L}_{m}+w_3\cdot\mathcal{L}_{h}, 
\label{Equation 7}
\end{equation}
where $\mathcal{L}_l$, $\mathcal{L}_m$, and $\mathcal{L}_h$ are individual loss for each hierarchical attention. $w$ is the adaptive weight (${\sum_{i=1}^{3}{w_i}=1}$) which determines the contribution of each hierarchical attention to the final prediction. 
Using the proposed loss function, our method effectively learns which features need to be suppressed or emphasized for image representation.
\section{EXPERIMENTS AND RESULTS}
This section details our experiments to evaluate the proposed method on several benchmark detasets.
\subsection{Implementation Setup}
% Following the implementation from \cite{arandjelovic2016netvlad,sarlin2019coarse,liu2019stochastic}, we use Pitts30k-training dataset \cite{arandjelovic2016netvlad} to train the proposed method. For each training query image $I_q$, we prepare for the positive $\{I^+_r\}$ and negative $\{I^-_r\}$ images correspondingly. We use all the potential positive images (images within 10 meters to the query input), and then we perform randomized negative data mining for the negative images (images are more than 25 meters to the query input). Eventually, we generate 120k image triplets in which 110k for training and 10k for validation.

In training, we use the margin m = 0.1, 30 epochs, learning rate 0.0001 which is halved in every 5 epochs, momentum 0.9, weight decay 0.001, and a batch size of 4 triplets. We use the Precision-Recall curve to evaluate the training performance\cite{arandjelovic2016netvlad}. The trained models which yield the best $recall@5$ on the validation set is used for testing. 
We utilize a grid search to find the best adaptive weights in training. Eventually, we have $w_1=0.1$, $w_2=0.4$, and $w_3=0.5$ respectively.
\subsection{Evaluation Datasets and Metrics}
We evaluate our method on two types of benchmarks: (i) Image retrieval datasets, which are Oxford 5k\cite{philbin2007object}, Paris 6k \cite{philbin2008lost}, and Holidays\cite{jegou2008hamming}. We employ the mean-Average-Precision (mAP) in  our evaluation; and (ii) Geo-localization datasets, which are Pitts250k-test\cite{torii2013visual}, Tokyo 24/7 \cite{torii201524}, TokyoTM-val\cite{arandjelovic2016netvlad}, and Sf-0\cite{chen2011city}. We  use  the  Precision-Recall  curve  to  test  the  performance of geo-localization.
%  \textbf{Image retrieval task.} Since  we cast  the  visual  localization  problem  as an  image  retrieval  task, we also evaluate our method on image retrieval benchmarks. Oxford 5k\cite{philbin2007object}, Paris 6k \cite{philbin2008lost}, and Holidays\cite{jegou2008hamming} are used to test the generalization of our method for image representations on image retrieval. We use the mean-Average-Precision (mAP) for evalution.
\subsection{Empirical Results}
To assess the benefits of the proposed method, we compare our method with the state-of-the-art methods, NetLAVD\cite{arandjelovic2016netvlad}, CRN\cite{jin2017learned}, and SuperPoint\cite{detone2018superpoint}, on geo-localization and image retrieval benchmarks. In order to have a fair comparison, we retrain all the methods with the same setup.
% \subsubsection{Image retrieval benchmarks}

\textbf{Image retrieval benchmarks.} To test the generalizability of our approach, our method is trained only on Pitts30k\cite{arandjelovic2016netvlad} without any fine-tuning on the image retrieval datasets. For Oxford 5k\cite{philbin2007object} and Paris 6k \cite{philbin2008lost}, we use both the full and cropped images; for Holidays\cite{jegou2008hamming}, we use original and rotated images. The results are displayed in Table \ref{retrival}. Our results set the state-of-the-art for compact image representations (256-D) on all three benchmarks. On all metrics, our margins consistently exceed the mAP of other methods by 1 to 5\%. For example, there are a 3.86\% improvements on Oxford 5k (full) than the next best method; and there are a 4\% improvements on Oxford 5k (crop) than the next best method. Our methods can be further improved by fine-tuning using the three image retrieval datasets.
\begin{table}[!ht]
    \centering
    \scriptsize
    \begin{tabular}{c|c|c|c|c|c|c}
    \toprule
    \multirow{2}{*}{Method}& \multicolumn{2}{c|}{Oxford 5K} & \multicolumn{2}{c|}{Paris 6k} & \multicolumn{2}{c}{Holidays}  \\
    \cline{2-7}
    \rule{0pt}{8pt} & full & crop & full & crop & orig & rot \\
    \midrule
    Ours & \textbf{67.81} & \textbf{69.52} & \textbf{75.10} & \textbf{78.29} & \textbf{84.82} & \textbf{88.41}  \\
    CRN & 63.95 & 65.52 & 72.88 & 75.85 & 83.19 & 87.30 \\
    NetVLAD & 63.09 & 65.33 & 72.53 & 75.67 & 82.67 & 86.83 \\
    SuperPoint & 63.14 & 65.50 & 72.83 & 75.10 & 82.92 & 86.90 \\
     \bottomrule
    \end{tabular}
    \caption{Results for compact image representations (256-D).}
    \label{retrival}
\end{table}

\begin{figure}[!ht]
  \centering
      \vskip -6pt
\includegraphics[width=8cm, trim={1.2cm 0 1.5cm 0},clip]{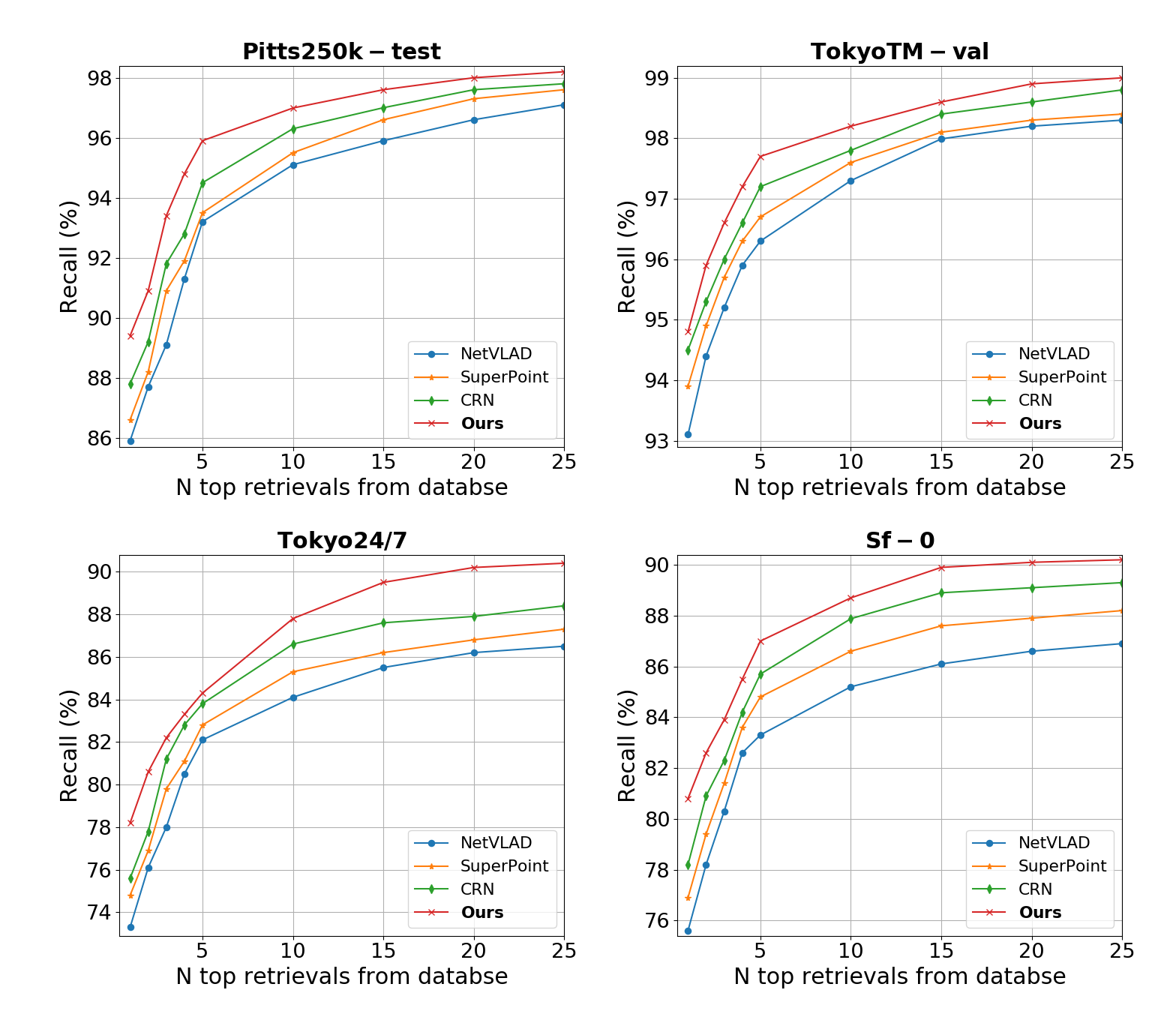}
  \caption{Comparison of recalls at $N$ top retrievals with the state-of-the-arts methods.}
  \label{sota}
        \vskip -6pt
\end{figure}

% \subsubsection{Geo-localization benchmarks} 
\textbf{Geo-localization benchmarks.} We report the  Precision-Recall plot for each method in Fig \ref{sota}. Our method outperforms other methods under different $recall@n$ thresholds on all benchmarks.  For qualitative analysis, we  use the heatmap to visualize the feature emphasis for localization using the input image (Fig. \ref{Figure 1}). The qualitative examples illuminate that our method can effectively exploit multi-scale features and demonstrate the capacity of having hierarchical attentions on landmarks with different scales and distances for geo-localization. In contrast, other methods mainly focus on large-scale landmarks for discriminative visual clues. Our method also focuses on the distinctive details of buildings while avoiding confusing objects such as pedestrians, vegetation, or vehicles which are hard for feature repeatability.

\indent Both quantitative and qualitative results advocate our hypothesis that we can leverage multi-scale features which are fused for hierarchical attention to produce discriminative yet compact image representations. 
\subsection{Adaptive Weight Analysis}
In training, we train a set of models using different combinations of weights with a change of 0.1. The adaptive weight $w_1$, $w_2$, and $w_3$ controls lower-level features (small scale), mid-level features (middle scale), and higher-level features (large scale) respectively. We report the best adaptive weight which produces the best $recall@5$ results for each benchmark in Table \ref{weight}. Pitts 250k-test dataset focuses on middle- and large-scale buildings. Thus, for Pitts 250k-test, $w_2$ and $w_3$ are much larger than $w_1$. TokyoTM dataset generally includes small-, middle-, and large-scale buildings which can be reflected from its best weights. For Tokyo 24/7, it has a similar adaptive weight as TokyoTM. Although Tokyo 24/7 has less small-scale buildings, it includes a lot of landmark details such as billboards, city lights, or traffic signs by the road. Sf-0 has a dominant $w_3$ as it mainly focuses on buildings with a large scale.
\begin{table}[!ht]
    \centering
      \vskip -6pt
    \scriptsize
    \begin{tabular}{c|c|c|c|c}
    \toprule
    \multirow{1}{*}{Method}& \multicolumn{1}{c|}{Pitts 250k-test} & \multicolumn{1}{c|}{TokyoTM-val} &
    \multicolumn{1}{c|}{Tokyo 24/7}&
    \multicolumn{1}{c}{Sf-0}  \\
    \midrule
    $w_1$ & 0.1 & 0.3 & 0.2 & 0.1  \\
     $w_2$ & 0.4 & 0.3 & 0.3 & 0.1  \\
    $w_3$ & 0.5 & 0.4 & 0.5 & 0.8  \\
     \bottomrule
    \end{tabular}
    \caption{Best adaptive weights for each benchmarks.}
    \label{weight}
      \vskip -12pt
\end{table}
\section{Conclusion}
 In this work, we introduce a hierarchical attention fusion network for geo-localization. We extract the multi-scale feature maps from a convolutional neural network (CNN) to perform hierarchical attention fusion for image representations. Since the hierarchical features are scale-sensitive, our method is robust to landmarks with different scales and distances. We evaluate our method extensively on the image retrieval benchmarks and the large-scale geo-localization benchmarks. Results indicate that our method is competitive with the latest state-of-the-art approaches.

\bibliographystyle{IEEEbib}
% \bibliography{strings,refs}
\bibliography{icassp.bib}

\begin{thebibliography}{10}

\bibitem{9206716}
D.~{Liu}, Y.~{Cui}, Z.~{Cao}, and Y.~{Chen},
\newblock ``A large-scale simulation dataset: Boost the detection accuracy for
  special weather conditions,''
\newblock in {\em 2020 IJCNN}, 2020, pp. 1--8.

\bibitem{9207265}
D.~{Liu}, Y.~{Cui}, Z.~{Cao}, and Y.~{Chen},
\newblock ``Indoor navigation for mobile agents: A multimodal vision fusion
  model,''
\newblock in {\em 2020 IJCNN}, 2020, pp. 1--8.

\bibitem{liu2020visual}
Dongfang Liu, Yiming Cui, Xiaolei Guo, Wei Ding, Baijian Yang, and Yingjie
  Chen,
\newblock ``Visual localization for autonomous driving: Mapping the accurate
  location in the city maze,''
\newblock {\em arXiv preprint arXiv:2008.05678}, 2020.

\bibitem{arandjelovic2016netvlad}
Relja Arandjelovic, Petr Gronat, Akihiko Torii, Tomas Pajdla, and Josef Sivic,
\newblock ``Netvlad: Cnn architecture for weakly supervised place
  recognition,''
\newblock in {\em Proceedings of the IEEE conference on computer vision and
  pattern recognition}, 2016, pp. 5297--5307.

\bibitem{salarian2018improved}
Mahdi Salarian, Nick Iliev, Ahmet~Enis Cetin, and Rashid Ansari,
\newblock ``Improved image-based localization using sfm and modified coordinate
  system transfer,''
\newblock {\em IEEE Transactions on Multimedia}, vol. 20, no. 12, pp.
  3298--3310, 2018.

\bibitem{sarlin2019coarse}
Paul-Edouard Sarlin, Cesar Cadena, Roland Siegwart, and Marcin Dymczyk,
\newblock ``From coarse to fine: Robust hierarchical localization at large
  scale,''
\newblock in {\em Proceedings of the IEEE Conference on Computer Vision and
  Pattern Recognition}, 2019, pp. 12716--12725.

\bibitem{jin2017learned}
Hyo Jin~Kim, Enrique Dunn, and Jan-Michael Frahm,
\newblock ``Learned contextual feature reweighting for image
  geo-localization,''
\newblock in {\em Proceedings of the IEEE Conference on Computer Vision and
  Pattern Recognition}, 2017, pp. 2136--2145.

\bibitem{detone2018superpoint}
Daniel DeTone, Tomasz Malisiewicz, and Andrew Rabinovich,
\newblock ``Superpoint: Self-supervised interest point detection and
  description,''
\newblock in {\em Proceedings of the IEEE Conference on Computer Vision and
  Pattern Recognition Workshops}, 2018, pp. 224--236.

\bibitem{simonyan2014very}
Karen Simonyan and Andrew Zisserman,
\newblock ``Very deep convolutional networks for large-scale image
  recognition,''
\newblock {\em arXiv preprint arXiv:1409.1556}, 2014.

\bibitem{LIU20201}
Dongfang Liu, Yiming Cui, Yingjie Chen, Jiyong Zhang, and Bin Fan,
\newblock ``Video object detection for autonomous driving: Motion-aid feature
  calibration,''
\newblock {\em Neurocomputing}, vol. 409, pp. 1 -- 11, 2020.

\bibitem{lin2017feature}
Tsung-Yi Lin, Piotr Doll{\'a}r, Ross Girshick, Kaiming He, Bharath Hariharan,
  and Serge Belongie,
\newblock ``Feature pyramid networks for object detection,''
\newblock in {\em Proceedings of the IEEE conference on computer vision and
  pattern recognition}, 2017, pp. 2117--2125.

\bibitem{huang2017densely}
Gao Huang, Zhuang Liu, Laurens Van Der~Maaten, and Kilian~Q Weinberger,
\newblock ``Densely connected convolutional networks,''
\newblock in {\em Proceedings of the IEEE conference on computer vision and
  pattern recognition}, 2017, pp. 4700--4708.

\bibitem{guo2020augfpn}
Chaoxu Guo, Bin Fan, Qian Zhang, Shiming Xiang, and Chunhong Pan,
\newblock ``Augfpn: Improving multi-scale feature learning for object
  detection,''
\newblock in {\em Proceedings of the IEEE/CVF Conference on Computer Vision and
  Pattern Recognition}, 2020, pp. 12595--12604.

\bibitem{dusmanu2019d2}
Mihai Dusmanu, Ignacio Rocco, Tomas Pajdla, Marc Pollefeys, Josef Sivic,
  Akihiko Torii, and Torsten Sattler,
\newblock ``D2-net: A trainable cnn for joint detection and description of
  local features,''
\newblock {\em arXiv preprint arXiv:1905.03561}, 2019.

\bibitem{philbin2007object}
James Philbin, Ondrej Chum, Michael Isard, Josef Sivic, and Andrew Zisserman,
\newblock ``Object retrieval with large vocabularies and fast spatial
  matching,''
\newblock in {\em 2007 IEEE conference on computer vision and pattern
  recognition}. IEEE, 2007, pp. 1--8.

\bibitem{philbin2008lost}
James Philbin, Ondrej Chum, Michael Isard, Josef Sivic, and Andrew Zisserman,
\newblock ``Lost in quantization: Improving particular object retrieval in
  large scale image databases,''
\newblock in {\em 2008 IEEE conference on computer vision and pattern
  recognition}. IEEE, 2008, pp. 1--8.

\bibitem{jegou2008hamming}
Herve Jegou, Matthijs Douze, and Cordelia Schmid,
\newblock ``Hamming embedding and weak geometric consistency for large scale
  image search,''
\newblock in {\em European conference on computer vision}. Springer, 2008, pp.
  304--317.

\bibitem{torii2013visual}
Akihiko Torii, Josef Sivic, Tomas Pajdla, and Masatoshi Okutomi,
\newblock ``Visual place recognition with repetitive structures,''
\newblock in {\em Proceedings of the IEEE conference on computer vision and
  pattern recognition}, 2013, pp. 883--890.

\bibitem{torii201524}
Akihiko Torii, Relja Arandjelovic, Josef Sivic, Masatoshi Okutomi, and Tomas
  Pajdla,
\newblock ``24/7 place recognition by view synthesis,''
\newblock in {\em Proceedings of the IEEE Conference on Computer Vision and
  Pattern Recognition}, 2015, pp. 1808--1817.

\bibitem{chen2011city}
David~M Chen, Georges Baatz, Kevin K{\"o}ser, Sam~S Tsai, Ramakrishna
  Vedantham, Timo Pylv{\"a}n{\"a}inen, Kimmo Roimela, Xin Chen, Jeff Bach, Marc
  Pollefeys, et~al.,
\newblock ``City-scale landmark identification on mobile devices,''
\newblock in {\em CVPR 2011}. IEEE, 2011, pp. 737--744.

\end{thebibliography}

\end{document}